\documentclass[conference]{IEEEtran}
\IEEEoverridecommandlockouts

\usepackage{cite}
\usepackage{amsmath,amssymb,amsfonts}
\usepackage{algorithmic}
\usepackage{graphicx}
\usepackage{textcomp}
\usepackage{xcolor}
\def\BibTeX{{\rm B\kern-.05em{\sc i\kern-.025em b}\kern-.08em
    T\kern-.1667em\lower.7ex\hbox{E}\kern-.125emX}}
\usepackage{enumitem}

\usepackage{url}

\usepackage{hyperref}

\begin{document}

\title{Benchmarking Human and Automated Prompting in the Segment Anything Model}

\author{
Jorge Quesada$^*$\thanks{$^*$Equal contribution}, Zoe Fowler$^*$, Mohammad Alotaibi, Mohit Prabhushankar, Ghassan AlRegib \\
  OLIVES at the Centre for Signal and Info. Processing\\
  Georgia Institute of Technology\\
  Atlanta, GA 30332, USA\\
  \texttt{\{jpacora3, zfowler3, malotaibi44, mohit.p, alregib\}@gatech.edu} \\

}

\onecolumn 

\begin{description}[labelindent=0cm,leftmargin=3cm,style=multiline]

\item[\textbf{Citation}]{J. Quesada$^*$, Z. Fowler$^*$, M. Alotaibi, M. Prabhushankar, and G. AlRegib, "Benchmarking Human and Automated Prompting in the Segment Anything Model", In IEEE International Conference on Big Data 2024, Washington DC, USA.}

\item[\textbf{Review}]{Date of acceptance: 26 October 2024}

\item[\textbf{Code}]{\url{https://github.com/olivesgatech/PointPrompt}}

\item[\textbf{Dataset}]{\url{https://zenodo.org/records/11580815}}

\item[\textbf{Bib}] {@ARTICLE\{Quesada2024Benchmarking,\\ 
author=\{J. Quesada and Z. Fowler and M. Alotaibi and M. Prabhushankar and G. AlRegib\},\\ 
journal=\{IEEE International Conference on Big Data\},\\ 
title=\{Benchmarking Human and Automated Prompting in the Segment Anything Model\}, \\ 
year=\{2024\},\\ 
keywords=\{foundation models, prompting, visual, segmentation\},\\ 
month=\{December\},\}
}


\item[\textbf{Copyright}]{\textcopyright 2024 IEEE. Personal use of this material is permitted. Permission from IEEE must be obtained for all other uses, in any current or future media, including reprinting/republishing this material for advertising or promotional purposes,
creating new collective works, for resale or redistribution to servers or lists, or reuse of any copyrighted component
of this work in other works. }

\item[\textbf{Contact}]{\href{mailto:olives.gatech@gmail.com}{olives.gatech@gmail.com} \\ \url{https://alregib.ece.gatech.edu/} \\ }
\end{description}

\thispagestyle{empty}
\newpage
\clearpage
\setcounter{page}{1}

\twocolumn

\maketitle

\begin{abstract}
The remarkable capabilities of the Segment Anything Model (SAM) for tackling image segmentation tasks in an intuitive and interactive manner has sparked interest in the design of effective visual prompts. Such interest has led to the creation of automated point prompt selection strategies, typically motivated from a feature extraction perspective.
However, there is still very little understanding of how appropriate these automated visual prompting strategies are, particularly when compared to humans, across diverse image domains. 
Additionally, the performance benefits of including such automated visual prompting strategies within the finetuning process of SAM also remains unexplored, as does the effect of interpretable factors like distance between the prompt points on segmentation performance.
To bridge these gaps, we leverage a recently released visual prompting dataset, PointPrompt, and introduce a number of benchmarking tasks that provide an array of opportunities to improve the understanding of the way human prompts differ from automated ones and what underlying factors make for effective visual prompts. 
We demonstrate that the resulting segmentation scores obtained by humans are approximately $29\%$ higher than those given by automated strategies and identify potential features that are indicative of prompting performance with $R^2$ scores over 0.5. Additionally, we demonstrate that performance when using automated methods can be improved by up to $68\%$ via a finetuning approach. 
Overall, our experiments not only showcase the existing gap between human prompts and automated methods, but also highlight potential avenues through which this gap can be leveraged to improve effective visual prompt design. Further details along with the dataset links and codes are available at \url{https://github.com/olivesgatech/PointPrompt}

\end{abstract}

\section{Introduction}
The way humans process and interact with our environment is often driven by different types of perceptual cues. For instance, we can visually recognize objects with quick localized glances without needing to gaze over the object entirely, due to our large degree of pre-existing visual knowledge. A long-standing goal in the machine learning community has been to replicate this level of efficiency and robustness in cue processing across different data modalities, leading to the emergence of foundation models \cite{bommasani2021opportunities}. The Segment Anything Model (SAM) \cite{kirillov2023segment} represents the newest advancement for foundation models in the computer vision domain, specializing in segmentation tasks. SAM allows users to provide prompts, such as bounding boxes or point prompts, promoting interactivity between SAM and the user.

While SAM has demonstrated impressive image segmentation results through its prompt-based approach, knowledge of what makes a prompt effective is still lacking. 
Existing literature has often explored the topic of prompt point selection from a feature extraction perspective, using automated methods like Shi-Tomasi or K-Medoids \cite{rajivc2023segment} or even vision saliency transformers to sample prompt points \cite{dai2023samaug}. However, understanding how effective these automated methods are, especially when compared to humans, remains largely unexplored. Utilizing and evaluating these automated approaches across diverse image domains is also lacking, as most are only applied to medical datasets and/or natural image and video datasets \cite{rajivc2023segment, dai2023samaug}. 

Additionally, SAM's overall effectiveness has been shown to vary when applied to more specialized out-of-distribution (OOD) domains, such as remote sensing or medical data. To this end, many prior approaches employ finetuning-based approaches to achieve enhanced segmentation performance in these specialized domains \cite{ma2024segment, wang2024samrs}. However, existing finetuning approaches rarely explore the extent of SAM's ability to handle a vast variety of prompts, particularly prompts that can be generated in an automated fashion. Instead, many prior works tend to finetune multiple components of SAM's architecture using images and only their corresponding bounding box prompts \cite{ma2024segment, li2023multi}. Hence, exploration into the impact of including a diverse range of prompts, such as those generated by humans or those generated via automated methods, within this finetuning process remains understudied.

\begin{figure*}
    \centering
    \includegraphics[width=0.9\linewidth]{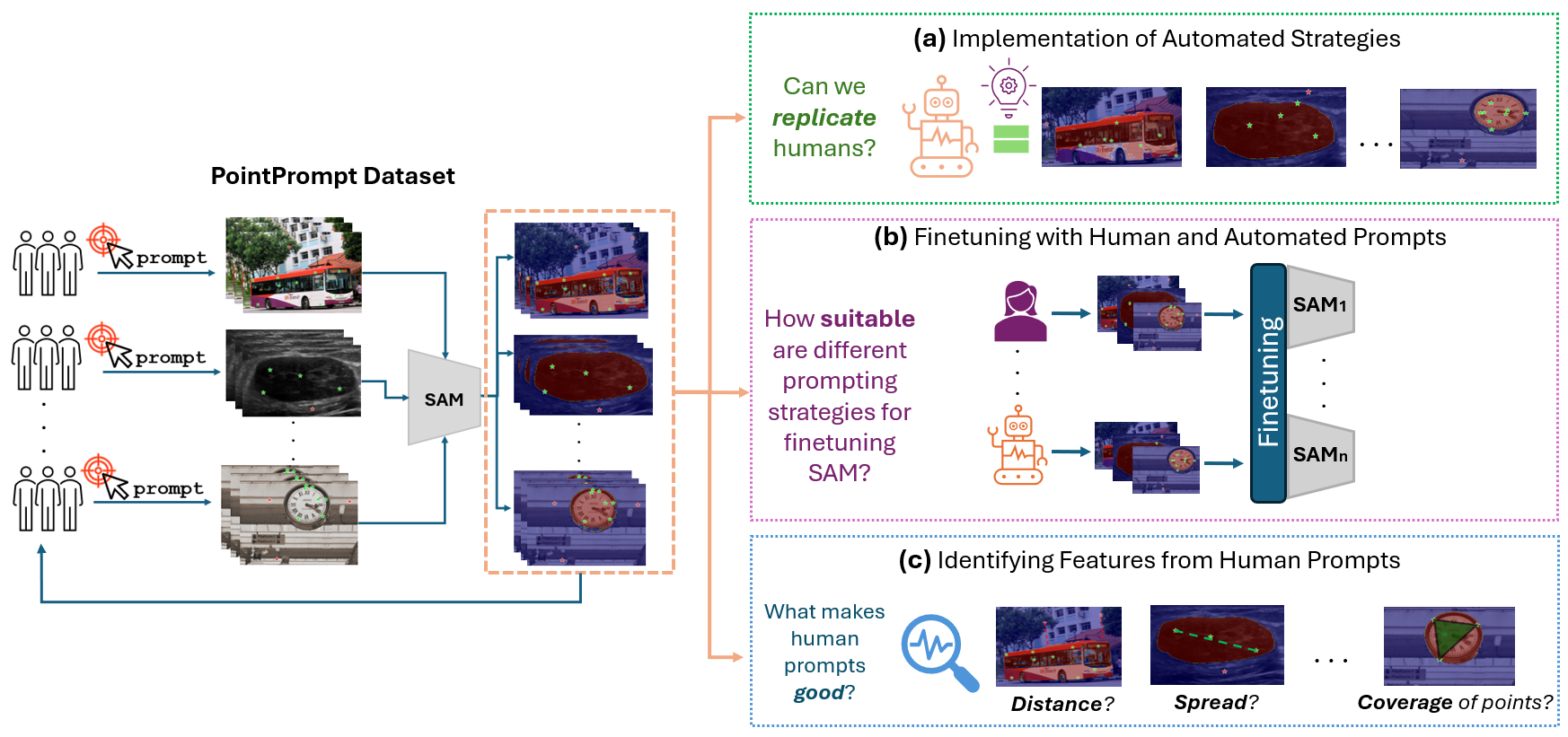}
    \caption{Our paper aims to \textbf{(a)} compare segmentation performance of automated point prompt generation strategies to human performance, \textbf{(b)} understand whether differences in human and automated performance can be mitigated through a finetuning-based approach, and \textbf{(c)} identify whether features derived from human-generated prompts can be indicative overall of segmentation performance to provide insights into characteristics of effective prompts.}
    \label{fig:overview}
\end{figure*}

While it is clear that both general-purpose prompt selection and OOD domain adaptation would benefit from a principled understanding of what factors constitute an effective prompt, prior research on identifying these factors is very limited. This can be largely attributed to the lack of open-source visual prompting data, which stands in contrast to the large volume of prompting data available in the language domain \cite{fan2018hierarchical, pister2024promptset, yu2022scaling}. In particular, part of the reason OOD adaptation frequently relies on model finetuning is that it is not clear how prompting strategies can be effectively adapted to handle these kind of inputs on base SAM. These knowledge gaps significantly limit the transferability of foundation models when effective prompting techniques are unknown and out-of-distribution domains are considered. 

The recent introduction of PointPrompt \cite{quesada2024pointprompt}, a comprehensive collection of prompting data across several image categories and multiple annotators per image, presents promising opportunities for a better understanding of the factors that underlie effective prompting strategies in the vision domain. In this work, we aim to understand how the prompts generated by humans differ from those generated by automated methods, exploring and quantifying these differences through three distinct benchmark tasks, summarized by Figure \ref{fig:overview}:

\begin{itemize}
    \item Task 1: We compare the performance attained by human prompting in this dataset against existing auto-prompting methods (as indicated in Figure \ref{fig:overview}.a) in the literature in a variety of experimental settings, clearly establishing and quantifying the existing gap between humans and automated methods. Additionally, we evaluate the degree of importance both inclusion and exclusion points have on segmentation performance through additional experiments.
    \item Task 2: We evaluate the extent to which finetuning SAM's prompt encoder to individual strategies can improve segmentation performance  and mitigate the performance differences between human and automated strategies (as shown in Figure \ref{fig:overview}.b,), albeit at the price of prompt curation and runtime costs.
    \item Task 3: We curate a series of intuitive features for each image/annotator pair (Figure \ref{fig:overview}.c), and evaluate the extent to which prompting performance can be decoded from these features. We analyze whether any features are particularly predictive of performance, and whether this can allow for the design of domain-specific prompting strategies based on the most significant features.
    
\end{itemize}

Our results on Task 1 establish the existence of a significant gap by which humans outperform automated methods, where automated methods on average experience a $29\%$ decline in segmentation performance as compared to humans. We also clearly showcase that inclusion points contribute much more to this gap than exclusion points, as utilizing human inclusion points in place of those generated by automated methods improves segmentation performance of the automated methods by approximately $36.3\%$ on average. Task 2 establishes that finetuning the prompt encoder is in fact an effective way to adapt SAM to particular strategies, achieving between $22$ to $68\%$ average increase over base SAM. Finally, we show in Task 3 that simple, interpretable features can be predictive of prompting performance in OOD domains (achieving an $R^2$ score above 0.5), and can therefore help guide effective prompt strategy design. These results not only showcase the differentiating advantage humans hold over automated methods in visual prompting, but highlight multiple avenues through which these differences can be studied and potentially leveraged in the development of better prompt engineering strategies.


\section{Related Works}
\subsection{SAM across Image Domains}
SAM has been deployed to a variety of image domains, though most prior work appears to target specific image domains. For example, \cite{ma2024segment} presents MedSAM, a medical image segmentation dataset acquired by finetuning SAM on medical data. \cite{huang2024segment} also tackles medical image segmentation using SAM. \cite{wang2024samrs} leverages SAM to construct a dataset for remote sensing segmentation. \cite{chen2024rsprompter} further explores SAM for remote sensing applications, aimed at developing effective prompts specifically for remote sensing images. Hence, prior work does not aim to understand SAM from a general perspective. Instead, the focus remains on adapting SAM to specialized domains, where solutions for what effective visual prompts might be are not widely generalizable. In this work, we address the lack of analyses of SAM across different image domains, accomplished by including natural, medical, seismic, and underwater images in our comparisons.

\subsection{PointPrompt Dataset}
Originally introduced in \cite{quesada2024pointprompt}, PointPrompt is a point-based visual prompting dataset built on SAM, collected by curating 6000 images to construct 16 image datasets from diverse domains. The image datasets were constructed by sampling 400 image/mask pairs from different publicly-available databases, as detailed below:
\begin{itemize}
    \item 9 datasets from the COCO \cite{lin2014microsoft} database, corresponding to the following classes: dog, cat, bird, clock, bus, baseball bat, cow, tie, stop sign.
    \item 2 datasets from the NDD20 \cite{trotter2020ndd20} database: dolphins above water and underwater.
    \item 3 medical datasets: Chest-X \cite{ALDHABYANI2020104863} (breast tumors), Kvasir-SEG \cite{jha2020kvasir} (polyp images) and ISIC \cite{gutman2016skin} (skin lesions).
    \item 2 seismic datasets from the F3 Facies database \cite{alaudah2019machine}, corresponding to the salt dome and chalk group categories.
\end{itemize}
 The general flow of annotation is depicted in Figure \ref{fig:overview} (left). A SAM-based prompting tool gives annotators the opportunity to add inclusion (inside object of interest, depicted in green) and exclusion (outside object of interest, depicted in red) points in an interactive click-based manner. Example masks generated by human annotators are shown in Figure \ref{fig:overview}.a.

The prompting data itself is a sequence corresponding to each individual image, including the prompt point coordinates, masks (generated by SAM for each set of prompt points), and Intersection Over Union (IoU) score between the SAM-produced and ground-truth (GT) masks at each step of the prompting process. 
The heterogeneity in both image domains and human prompting strategies in this dataset constitutes a rich test bed for addressing many of the existing knowledge gaps in the visual prompting domain.

\subsection{Prompt Point Selection Strategies}
\label{point-selection}
SAM easily allows the user to prompt a given image with inclusion and exclusion point prompts. SAM uses inclusion points to denote the object that should be included in the generated mask, while exclusion points serve to show SAM objects that should not be included. However, both the number and location of these points can have a drastic impact on the output mask, as described in \cite{kirillov2023segment} where single point prompts generate potential ambiguities in the output mask. Furthermore, prior literature has observed that automatically generating prompting points can lead to improved segmentation results \cite{rajivc2023segment, dai2023samaug}. 
From prior literature, we detail a list of six automated point selection strategies that have been used in conjunction with SAM:
\begin{enumerate}
    \item \textbf{Random Sampling} samples prompt points from the provided mask randomly \cite{dai2023samaug, rajivc2023segment}.
    \item \textbf{K-Medoids Sampling} samples prompt points by calculating the cluster centers of the provided mask using K-Medoids clustering, as implemented in \cite{rajivc2023segment, park2009simple}.
    \item \textbf{Shi-Tomasi Sampling} samples prompt points that correspond to the corners of the image covered by the provided mask, as implemented in \cite{rajivc2023segment, shi1994good}.
    \item \textbf{Entropy Sampling} randomly samples a singular point from the provided mask, selecting the next points based on those that maximize the difference in entropy with respect to this original point \cite{dai2023samaug}. 
    \item \textbf{Maximum Distance Sampling} randomly samples a singular point from the provided mask, selecting the next points based on those that maximize the Euclidean distance with respect to this original point as described in \cite{dai2023samaug}.
    \item \textbf{Saliency Sampling} randomly samples a point from the provided mask, using SAM to generate an initial mask. The region of the image captured by this initial mask is then fed into a vision saliency transformer \cite{liu2021visual}, marking which areas of the image are the most salient. Prompt points are then randomly selected from the more salient regions \cite{dai2023samaug}.
    
\end{enumerate}

All prior point selection methods mentioned exclusively focus on sampling from an \textit{inclusion} perspective, where the provided mask is the ground truth mask. In this work, however, we also sample exclusion points. To the best of our knowledge, there are no prior works that focus on the sampling of exclusion points, and this work is the first to consider doing so.

\section{Benchmarking Tasks}

\subsection{Benchmarking Tasks}
\subsubsection{Automated Prompt Point Selection}
\label{method-auto}
\begin{figure*}[h!]
    \centering
    \includegraphics[width=0.9\linewidth]{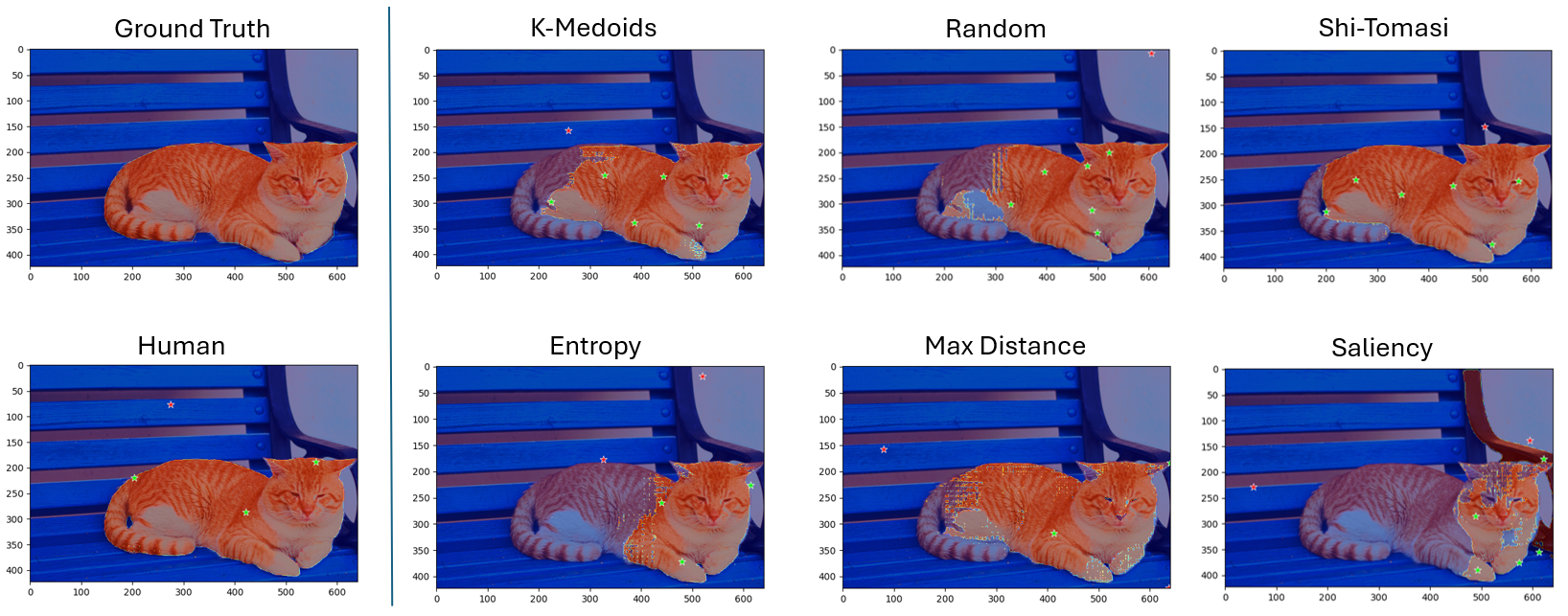}
    \caption{Mask outputs when using automated strategies versus a human annotator's output mask}
    \label{fig:mask-outputs}
\end{figure*}

Understanding the effectiveness of automated prompt point selection strategies, particularly in comparison to points generated by humans, remains largely understudied.
Therefore, in this section, we discuss experimental setups for closing this gap. In particular, we detail the procedure for automatically generating both inclusion and exclusion prompt points. Specifically, we aim to implement the six automated strategies for generating prompt points from Section \ref{point-selection}, which composes $P = $ \{random, K-Medoids, Shi-Tomasi, entropy, maximum distance, and saliency\} (see Figure \ref{fig:mask-outputs} for visual examples of these strategies). We then later explore the impact of using either human inclusion or exclusion points in conjunction with the different automated strategies. Thus, our experimental setups for this section aim to assess the impact of utilizing (1) the same automated strategy for selection of inclusion and exclusion points and (2) both automated and human prompt points jointly.

SAM is composed of an image encoder, a mask decoder, and a prompt encoder. Hence, for these experiments, we take the $j$-th image dataset for $j \in [0,N]$, where $N$ corresponds to the total number of image datasets. All images from dataset $j$ are then fed into the fixed image encoder.
Then, for automatically generating prompt points, we sample \textit{inclusion points} using each of the six sampling strategies $p \in P$ described in Section \ref{task1}, where all strategies except the Saliency strategy sample points from the ground truth mask (as commonly done in prior literature \cite{dai2023samaug, rajivc2023segment}). This results in the creation of $I_{j,p}$, which stores the inclusion points for a particular automated strategy $p$. Similarly, we sample \textit{exclusion points} using the same six sampling strategies $p \in P$, where the background is instead considered instead of the ground truth mask for sampling points. This process results in $E_{j,p}$, which stores the exclusion points for automated strategy $p$ on image dataset $j$.
The inclusion and exclusion points selected are then fed into the prompt encoder of SAM. This leads to the creation of an array $IoU_{j,p}$, which stores all IoU scores for the images corresponding to dataset $j$ for an automated strategy $p$. 

\begin{figure}[h!]
    \centering
    \includegraphics[width=\linewidth]{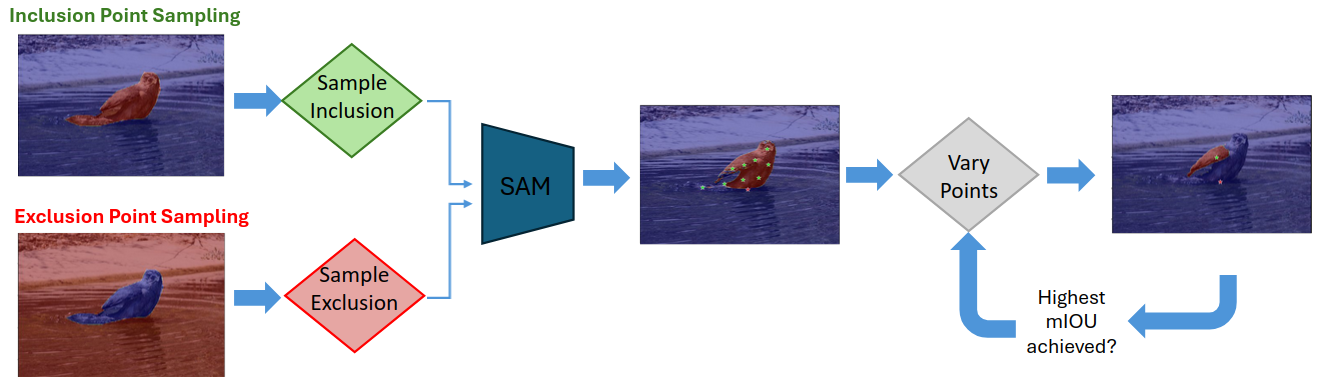}
    \caption{Framework for determining optimal number of inclusion and exclusion points, where points are originally sampled based on the average number of inclusion/exclusion points used by the human annotators for a specific image category}
    \label{fig:automated-flow}
\end{figure}

Inclusion and exclusion points are first sampled in an amount equivalent to the average number sampled by human annotators for a particular dataset in PointPrompt as reported in \cite{quesada2024pointprompt}. Afterwards, we assess the amount of inclusion and exclusion points necessary to give the highest mean IoU (mIoU) score (given by averaging the scores in $IoU_{j,p}$) for a particular image dataset $j$. To accomplish this, we follow Figure \ref{fig:automated-flow}. In Figure \ref{fig:automated-flow}, we alternate the number of inclusion/exclusion points considered, recording those that overall gave the highest mIoU score. Hence, $IoU_{j,p}$ becomes the array of IoU scores achieved when using the `best' number of inclusion and exclusion points.
The final mIoU score of image dataset $j$ using a particular automated strategy $p$ is then derived by averaging the scores in $IoU_{j,p}$.

\subsubsection{Finetuning}
Another seldom explored question regarding SAM's capabilities is to what extent it can be adapted to handle a variety of prompting strategies, both human and automated.
To this end, we compare the automated methods implemented in the previous task with human-generated prompts in a finetuning scenario, where SAM is finetuned on each prompting method and the resulting models are compared.

To achieve this, we freeze the image encoder and mask decoder components of SAM, training only the prompt encoder. For each strategy, a training set of 300 samples (comprising images and their corresponding prompts) from each dataset is used as input to train the prompt encoder. A subset of 10\% from the training set is used to validate the model after each epoch, and if there is no improvement in the validation loss for five epochs, the training is stopped. The final models are then tested on a held-out set of 100 samples. We use the Dice loss and Adam optimizer \cite{kingma2017adammethodstochasticoptimization} in our experiments, with an initial learning rate of 0.001.

\subsubsection{Feature extraction and performance decoding}
Another important requirement both for automated prompt generation and the study of the differences between human and automated prompts is understanding whether there are underlying factors in a prompting strategy that lead to good segmentation performance and are human-interpretable. If possible, these factors could provide a way to build efficient and principled prompt generation approaches for individual domains. To this end, we curate 26 interpretable features for each image/annotator pair, which can be broadly categorized as data-level or prompt-level features:
\begin{itemize}
    \item \textbf{Data-level features}: These are features that describe the image or object of interest in each image, without considering the way in which the image is annotated.
    \begin{itemize}
        \item Mask size: Percentage of object pixels in an image
        \item Connected Component density ($CCD$): Number of binary objects in the GT mask, obtained by connected component analysis. 
        \item Object density ($OD$): Number of object instances in image (manually annotated).
        \item Merged, Split and Compact: Binary variables that are a function of $OD$ and $CCD$. They indicate whether a given GT mask is merging multiple objects ($OD>CCD$), splitting single objects ($OD<CCD$) or is compactly encoding the objects ($OD=CCD$).
        \item GCLM features: A number of features based on the gray level co-ocurrence matrix (GCLM, \cite{haralick1973textural}) were collected: contrast, dissimilarity, homogeneity, energy, correlation and angular second moment (ASM). These features were collected at both object and background level for all images.
    \end{itemize}
    \item \textbf{Prompt-level features}: These are features that describe the prompt points generated by the annotators, and therefore are meant to provide a window to study the way in which the annotators approach the prompting task.
    \begin{itemize}
        \item Prompt coverage: Metric for how much of the object area is `covered' by the prompts, achieved by taking the convex hull of the inclusion points intersected with the GT and dividing its area by the area of the object.
        \item Coverage efficiency: Prompt coverage divided by number of inclusion points
        \item Inclusion and exclusion maximum spread: Maximum intra-set distance for the sets of inclusion and exclusion points. Normalized by object diameter for inclusion and by image diagonal size for exclusion.
        \item Average minimum distance: Calculates distances to the nearest point for each inclusion point, and takes the average across these distances.
        \item Inclusion/exclusion distance: Average distance between the set of inclusion and exclusion points, as measured by the Chamfer distance \cite{besl1992method}
        \item Inclusion and exclusion margin: Average distance to object boundary for the sets of inclusion and exclusion points. Normalized by object diameter for inclusion and by image diagonal size for exclusion.
    \end{itemize}
\end{itemize}
A visual example of inclusion/exclusion distance, inclusion maximum spread, and coverage is shown in Figure \ref{fig:overview}.c.

We explore the role data and prompt-level features play into segmentation performance by performing both correlation-level analyses and decoding experiments. The correlation analysis allows us to measure individual relationships between these features and the target score, and decoding experiments provide an understanding of how the interplay between these features affects performance. We restrict the latter experiments to simple decoding strategies like linear and polynomial regression instead of model-based approaches, given that we want to measure the existence of deterministic relationships between the features and the target score, rather than treating the decoding operation as a black-box system.

We perform both analyses by combining the 198,590 sets of prompts in PointPrompt and the 36,000 sets of prompts form the implementation of Task 1. We extract both data and prompt-level features for these 234,590 prompt sets to perform both the correlation and performance decoding analyses. For the latter, the prompts corresponding to each dataset are split into a train and test set following an 80/20 split ratio, and a linear or polynomial regression is performed on the train set and evaluated on the test set to evaluate the extent to which the features are predictive of performance.

\section{Results}
\subsection{Benchmarking Automated Strategy Performance}
\label{task1}
\subsubsection{Same Strategy Sampling}
\label{task1.1}

\begin{table*}[ht]
\caption{\textbf{Same strategy sampling:} Resulting mIoU (± standard deviation) across image datasets for automated baselines, compared to human mIoUs.}
\label{tab:strategies-best}
\resizebox{\textwidth}{!}{%
\begin{tabular}{cccccccc}
\hline
\textbf{Category} & \textbf{Human mIoU}    & \textbf{Random mIoU} & \textbf{Saliency mIoU} & \textbf{K-Medoids mIoU}  & \textbf{Entropy mIoU} & \textbf{Maximum Dist mIoU} & \textbf{Shi-Tomasi mIoU} \\
\hline
\hline
Baseball bat      & \textbf{0.747 ± 0.152} & 0.684 ± 0.198      & 0.422 ± 0.323          & 0.724 ± 0.172          & 0.653 ± 0.217         & 0.632 ± 0.249           & 0.701 ± 0.178            \\
Bird              & \textbf{0.677 ± 0.231} & 0.615 ± 0.222      & 0.308 ± 0.300          & 0.645 ± 0.212          & 0.483 ± 0.267         & 0.456 ± 0.296           & 0.620 ± 0.216            \\
Bus               & \textbf{0.803 ± 0.144} & 0.593 ± 0.196      & 0.158 ± 0.190          & 0.636 ± 0.172          & 0.359 ± 0.260         & 0.289 ± 0.277           & 0.548 ± 0.204            \\
Cat               & \textbf{0.887 ± 0.079} & 0.771 ± 0.149      & 0.487 ± 0.329          & 0.825 ± 0.108          & 0.583 ± 0.284         & 0.508 ± 0.342           & 0.795 ± 0.140            \\
Clock             & \textbf{0.814 ± 0.181} & 0.745 ± 0.209      & 0.432 ± 0.338          & 0.735 ± 0.205          & 0.680 ± 0.262         & 0.692 ± 0.265           & 0.715 ± 0.223            \\
Cow               & \textbf{0.808 ± 0.130} & 0.675 ± 0.189      & 0.321 ± 0.294          & 0.660 ± 0.185          & 0.423 ± 0.280         & 0.343 ± 0.314           & 0.646 ± 0.187            \\
Dog               & \textbf{0.848 ± 0.102} & 0.742 ± 0.166      & 0.436 ± 0.320          & 0.780 ± 0.135          & 0.544 ± 0.271         & 0.523 ± 0.317           & 0.745 ± 0.161            \\
Tie               & \textbf{0.700 ± 0.276} & 0.627 ± 0.292      & 0.368 ± 0.352          & 0.649 ± 0.289          & 0.569 ± 0.309         & 0.542 ± 0.337           & 0.646 ± 0.283            \\
Stop sign         & \textbf{0.886 ± 0.132} & 0.839 ± 0.169      & 0.550 ± 0.384          & 0.848 ± 0.158          & 0.773 ± 0.248         & 0.726 ± 0.314           & 0.640 ± 0.273            \\
Dolphin above     & \textbf{0.732 ± 0.100} & 0.642 ± 0.112      & 0.472 ± 0.249          & 0.655 ± 0.106          & 0.553 ± 0.176         & 0.543 ± 0.209           & 0.661 ± 0.099            \\
Dolphin below     & \textbf{0.831 ± 0.075} & 0.670 ± 0.135      & 0.391 ± 0.303          & 0.714 ± 0.114          & 0.473 ± 0.254         & 0.430 ± 0.315           & 0.681 ± 0.108            \\
Polyp             & \textbf{0.794 ± 0.145} & 0.747 ± 0.164      & 0.547 ± 0.326          & 0.757 ± 0.151          & 0.634 ± 0.290         & 0.415 ± 0.354           & 0.637 ± 0.246            \\
Skin              & 0.593 ± 0.195          & 0.593 ± 0.196      & 0.375 ± 0.295          & \textbf{0.626 ± 0.154} & 0.452 ± 0.283         & 0.395 ± 0.317           & 0.515 ± 0.208            \\
Breast            & \textbf{0.750 ± 0.126} & 0.621 ± 0.237      & 0.438 ± 0.336          & 0.674 ± 0.194          & 0.566 ± 0.286         & 0.532 ± 0.335           & 0.592 ± 0.305            \\
Salt dome         & \textbf{0.844 ± 0.096} & 0.513 ± 0.141      & 0.273 ± 0.177          & 0.588 ± 0.101          & 0.347 ± 0.197         & 0.306 ± 0.162           & 0.564 ± 0.111            \\
Chalk group       & \textbf{0.714 ± 0.101} & 0.409 ± 0.125      & 0.237 ± 0.135          & 0.441 ± 0.118          & 0.308 ± 0.143         & 0.299 ± 0.129           & 0.425 ± 0.121
\\
\hline       
\end{tabular}%
}
\end{table*}
We first benchmark how well automated strategies perform when compared to humans. Specifically, for this setup, prompt points for the automated strategies are selected as described in Section \ref{method-auto}, where each strategy $p \in P$ is used to sample \textit{both} inclusion and exclusion points. Example mask outputs for all six strategies are demonstrated in Figure \ref{fig:mask-outputs}, with a comparison against a human annotator's output mask and the ground truth. Quantitative results for this experimental setup are reported in Table \ref{tab:strategies-best}, where the highest mIoU achieved per image category across the different point sampling strategies is bolded. As demonstrated in Table \ref{tab:strategies-best}, the human annotators achieve the highest mIoU on all image datasets except the Skin dataset.  Additionally, there are many image datasets where the humans significantly outperform the automated strategies, such as in the seismic domain (Salt dome and Chalk group).

Humans also design more effective prompts for some natural image datasets, such as the Bus and Cow dataset. These results suggest that humans are overall more effective prompters than automated methods, as we can use the results in Table \ref{tab:strategies-best} to see that automated methods on average experience a $29\%$ decline in performance when compared to humans. These results also indicate that for certain domains, such as seismic, the way that human annotators prompt data is inherently different, as all automated strategies obtain low mIoU scores.

\subsubsection{Human Exclusion Sampling}
\label{task1.2}
For this setup, we utilize the same optimal inclusion points $I_{j,p}$ determined in Section \ref{task1.1} for each of the six sampling strategies across the different image datasets. Instead of using $E_{j,p}$, however, \textit{we use the exact exclusion points determined by the human annotators}. Hence, we aim to determine the impact human exclusion points have on the resulting mask (when only inclusion points are sampled using automated strategies). As there are multiple annotators per dataset, the IoU for a specific image in an image dataset is determined by averaging the IoUs that result from using the different human annotator exclusion points. The final mean IoU for an image dataset is then the average of the mean IoU for each image across different annotators.
\begin{table*}[ht]
\caption{\textbf{Human Exclusion Sampling:} mIoU Results and Percent Change from Table \ref{tab:strategies-best}}
\label{tab:human-exclusion2}
\resizebox{\textwidth}{!}{%
\begin{tabular}{cccccccc}
\hline
\textbf{Category} & \textbf{Human mIoU}    & \textbf{Random mIoU} & \textbf{Saliency mIoU} & \textbf{K-Medoids mIoU}  & \textbf{Entropy mIoU} & \textbf{Maximum Dist mIoU} & \textbf{Shi-Tomasi mIoU} \\
\hline\hline
Baseball bat  & \textbf{0.747 ± 0.152} & 0.688 (\textcolor{green}{$\uparrow$} 0.58\%)   & 0.426 (\textcolor{green}{$\uparrow$} 0.95\%)   & 0.731 (\textcolor{green}{$\uparrow$} 0.97\%)  & 0.654 (\textcolor{green}{$\uparrow$} 0.15\%)   & 0.627 (\textcolor{red}{$\downarrow$} -0.79\%)  & 0.709 (\textcolor{green}{$\uparrow$} 1.14\%)   \\
Bird          & \textbf{0.677 ± 0.231} & 0.619 (\textcolor{green}{$\uparrow$} 0.65\%)   & 0.315 (\textcolor{green}{$\uparrow$} 2.27\%)   & 0.648 (\textcolor{green}{$\uparrow$} 0.47\%)  & 0.484 (\textcolor{green}{$\uparrow$} 0.21\%)   & 0.438 (\textcolor{red}{$\downarrow$} -3.95\%)  & 0.619 (\textcolor{red}{$\downarrow$} -0.16\%)  \\
Bus           & \textbf{0.803 ± 0.144} & 0.576 (\textcolor{red}{$\downarrow$} -2.87\%)  & 0.157 (\textcolor{red}{$\downarrow$} -0.63\%)  & 0.629 (\textcolor{red}{$\downarrow$} -1.10\%) & 0.356 (\textcolor{red}{$\downarrow$} -0.84\%)  & 0.281 (\textcolor{red}{$\downarrow$} -2.77\%)  & 0.53 (\textcolor{red}{$\downarrow$} -3.28\%)   \\
Cat           & \textbf{0.887 ± 0.079} & 0.767 (\textcolor{red}{$\downarrow$} -0.52\%)  & 0.504 (\textcolor{green}{$\uparrow$} 3.49\%)   & 0.824 (\textcolor{red}{$\downarrow$} -0.12\%) & 0.571 (\textcolor{red}{$\downarrow$} -2.06\%)  & 0.469 (\textcolor{red}{$\downarrow$} -7.68\%)  & 0.785 (\textcolor{red}{$\downarrow$} -1.26\%)  \\
Clock         & \textbf{0.814 ± 0.181} & 0.764 (\textcolor{green}{$\uparrow$} 2.55\%)   & 0.439 (\textcolor{green}{$\uparrow$} 1.62\%)   & 0.759 (\textcolor{green}{$\uparrow$} 3.27\%)  & 0.698 (\textcolor{green}{$\uparrow$} 2.65\%)   & 0.688 (\textcolor{red}{$\downarrow$} -0.58\%)  & 0.73 (\textcolor{green}{$\uparrow$} 2.10\%)    \\
Cow           & \textbf{0.808 ± 0.130} & 0.672 (\textcolor{red}{$\downarrow$} -0.44\%)  & 0.322 (\textcolor{green}{$\uparrow$} 0.31\%)   & 0.656 (\textcolor{red}{$\downarrow$} -0.61\%) & 0.418 (\textcolor{red}{$\downarrow$} -1.18\%)  & 0.308 (\textcolor{red}{$\downarrow$} -10.20\%) & 0.634 (\textcolor{red}{$\downarrow$} -1.86\%)  \\
Dog           & \textbf{0.848 ± 0.102} & 0.742 ( 0.00\%)                              & 0.445 (\textcolor{green}{$\uparrow$} 2.06\%)   & 0.786 (\textcolor{green}{$\uparrow$} 0.77\%)  & 0.545 (\textcolor{green}{$\uparrow$} 0.18\%)   & 0.508 (\textcolor{red}{$\downarrow$} -2.87\%)  & 0.74 (\textcolor{red}{$\downarrow$} -0.67\%)   \\
Tie           & \textbf{0.700 ± 0.276} & 0.638 (\textcolor{green}{$\uparrow$} 1.75\%)   & 0.395 (\textcolor{green}{$\uparrow$} 7.34\%)   & 0.668 (\textcolor{green}{$\uparrow$} 2.93\%)  & 0.578 (\textcolor{green}{$\uparrow$} 1.58\%)   & 0.537 (\textcolor{red}{$\downarrow$} -0.92\%)  & 0.663 (\textcolor{green}{$\uparrow$} 2.63\%)   \\
Stop sign     & \textbf{0.886 ± 0.132} & 0.84 (\textcolor{green}{$\uparrow$} 0.12\%)    & 0.555 (\textcolor{green}{$\uparrow$} 0.91\%)   & 0.853 (\textcolor{green}{$\uparrow$} 0.59\%)  & 0.767 (\textcolor{red}{$\downarrow$} -0.78\%)  & 0.718 (\textcolor{red}{$\downarrow$} -1.10\%)  & 0.622 (\textcolor{red}{$\downarrow$} -2.81\%)  \\
Dolphin above & \textbf{0.732 ± 0.100} & 0.642 ( 0.00\%)                              & 0.49 (\textcolor{green}{$\uparrow$} 3.81\%)    & 0.658 (\textcolor{green}{$\uparrow$} 0.46\%)  & 0.551 (\textcolor{red}{$\downarrow$} -0.36\%)  & 0.532 (\textcolor{red}{$\downarrow$} -2.03\%)  & 0.657 (\textcolor{red}{$\downarrow$} -0.61\%)  \\
Dolphin below & \textbf{0.831 ± 0.075} & 0.679 (\textcolor{green}{$\uparrow$} 1.34\%)   & 0.398 (\textcolor{green}{$\uparrow$} 1.79\%)   & 0.724 (\textcolor{green}{$\uparrow$} 1.40\%)  & 0.459 (\textcolor{red}{$\downarrow$} -2.96\%)  & 0.403 (\textcolor{red}{$\downarrow$} -6.28\%)  & 0.685 (\textcolor{green}{$\uparrow$} 0.59\%)   \\
Polyp         & \textbf{0.794 ± 0.145} & 0.745 (\textcolor{red}{$\downarrow$} -0.27\%)  & 0.55 (\textcolor{green}{$\uparrow$} 0.55\%)    & 0.762 (\textcolor{green}{$\uparrow$} 0.66\%)  & 0.617 (\textcolor{red}{$\downarrow$} -2.68\%)  & 0.437 (\textcolor{green}{$\uparrow$} 5.30\%)   & 0.601 (\textcolor{red}{$\downarrow$} -5.65\%)  \\
Skin          & 0.593 ± 0.195          & 0.573 (\textcolor{red}{$\downarrow$} -3.37\%)  & 0.377 (\textcolor{green}{$\uparrow$} 0.53\%)   &\textbf{ 0.629} (\textcolor{green}{$\uparrow$} 0.48\%)  & 0.446 (\textcolor{red}{$\downarrow$} -1.33\%)  & 0.384 (\textcolor{red}{$\downarrow$} -2.78\%)  & 0.483 (\textcolor{red}{$\downarrow$} -6.21\%)  \\
Breast        & \textbf{0.750 ± 0.126} & 0.619 (\textcolor{red}{$\downarrow$} -0.32\%)  & 0.455 (\textcolor{green}{$\uparrow$} 3.88\%)   & 0.679 (\textcolor{green}{$\uparrow$} 0.74\%)  & 0.576 (\textcolor{green}{$\uparrow$} 1.77\%)   & 0.523 (\textcolor{red}{$\downarrow$} -1.69\%)  & 0.579 (\textcolor{red}{$\downarrow$} -2.20\%)  \\
Salt dome     & \textbf{0.844 ± 0.096} & 0.478 (\textcolor{red}{$\downarrow$} -6.82\%)  & 0.234 (\textcolor{red}{$\downarrow$} -14.29\%) & 0.568 (\textcolor{red}{$\downarrow$} -3.40\%) & 0.307 (\textcolor{red}{$\downarrow$} -11.53\%) & 0.276 (\textcolor{red}{$\downarrow$} -9.80\%)  & 0.516 (\textcolor{red}{$\downarrow$} -8.51\%)  \\
Chalk group   & \textbf{0.714 ± 0.101} & 0.349 (\textcolor{red}{$\downarrow$} -14.67\%) & 0.129 (\textcolor{red}{$\downarrow$} -45.57\%) & 0.399 (\textcolor{red}{$\downarrow$} -9.52\%) & 0.22 (\textcolor{red}{$\downarrow$} -28.57\%)  & 0.169 (\textcolor{red}{$\downarrow$} -43.48\%) & 0.364 (\textcolor{red}{$\downarrow$} -14.35\%)\\
\hline
\end{tabular}%
}
\end{table*}

Results for this experiment are reported in Table \ref{tab:human-exclusion2}, where the highest mIoU score per image category is bolded.
Table \ref{tab:human-exclusion2} also reports the percent change from Table \ref{tab:strategies-best}, denoted by arrows.
Table \ref{tab:human-exclusion2} demonstrates that when human exclusion points are used (instead of sampled via an automated strategy), mIoU performance hardly improves across the different image datasets. In fact, when computing the overall average percent change experienced across all strategies, we observe that performance deteriorates from the results reported in Table \ref{tab:strategies-best} by $2.43\%$ on average. This decline in mIoU is prevalent throughout the table, shown clearly with strategies like Maximum Distance. The Saliency and K-Medoids strategy, however, tends to improve, though the performance enhancements are marginal. Overall, Table \ref{tab:human-exclusion2} indicates that the exclusion point sampling process may not be as relevant to obtaining high mIoU scores.

\subsubsection{Human Inclusion Sampling}
\label{task1.3}
Similar to the previous setup, we now utilize \textit{the exact inclusion points determined by the human annotators} for this setup. We utilize $E_{j,p}$ for the exclusion points as determined in Section \ref{task1.1} for each automated strategy. 
In other words, this setup focuses on determining the impact of human inclusion points when exclusion points are chosen in the automated fashion demonstrated in Table \ref{tab:strategies-best}. As done in Section \ref{task1.2},
the IoU for a specific image is determined by averaging the individual IoUs that result from using the different human annotator inclusion points. The final mean IoU for an image dataset is then the average of the IoUs for each image across different annotators.

\begin{table*}[ht]
\caption{\textbf{Human Inclusion Sampling:} mIoU Results and Percent Change from Table \ref{tab:strategies-best}}
\label{tab:human-inclusion-arrows}
\resizebox{\textwidth}{!}{%
\begin{tabular}{cccccccc}
\hline
\textbf{Category} & \textbf{Human mIoU}    & \textbf{Random mIoU} & \textbf{Saliency mIoU} & \textbf{K-Medoids mIoU}  & \textbf{Entropy mIoU} & \textbf{Maximum Dist mIoU} & \textbf{Shi-Tomasi mIoU} \\
\hline\hline
Baseball bat  & \textbf{0.747 ± 0.152} & 0.729 (\textcolor{green}{$\uparrow$} 6.58\%)  & 0.729 (\textcolor{green}{$\uparrow$} 72.75\%)  & 0.731 (\textcolor{green}{$\uparrow$} 0.97\%)   & 0.731 (\textcolor{green}{$\uparrow$} 11.94\%)  & 0.73 (\textcolor{green}{$\uparrow$} 15.51\%)   & 0.73 (\textcolor{green}{$\uparrow$} 4.14\%)    \\
Bird          & \textbf{0.677 ± 0.231} & 0.672 (\textcolor{green}{$\uparrow$} 9.27\%)  & 0.668 (\textcolor{green}{$\uparrow$} 116.88\%) & 0.671 (\textcolor{green}{$\uparrow$} 4.03\%)   & 0.672 (\textcolor{green}{$\uparrow$} 39.13\%)  & 0.673 (\textcolor{green}{$\uparrow$} 47.59\%)  & 0.67 (\textcolor{green}{$\uparrow$} 8.06\%)    \\
Bus           & \textbf{0.803 ± 0.144} & 0.784 (\textcolor{green}{$\uparrow$} 32.21\%) & 0.738 (\textcolor{green}{$\uparrow$} 367.09\%) & 0.779 (\textcolor{green}{$\uparrow$} 22.48\%)  & 0.764 (\textcolor{green}{$\uparrow$} 112.81\%) & 0.779 (\textcolor{green}{$\uparrow$} 169.55\%) & 0.787 (\textcolor{green}{$\uparrow$} 43.61\%)  \\
Cat           & \textbf{0.887 ± 0.079} & 0.874 (\textcolor{green}{$\uparrow$} 13.36\%) & 0.841 (\textcolor{green}{$\uparrow$} 72.69\%)  & 0.872 (\textcolor{green}{$\uparrow$} 5.70\%)   & 0.87 (\textcolor{green}{$\uparrow$} 49.23\%)   & 0.87 (\textcolor{green}{$\uparrow$} 71.26\%)   & 0.876 (\textcolor{green}{$\uparrow$} 10.19\%)  \\
Clock         & \textbf{0.814 ± 0.181} & 0.758 (\textcolor{green}{$\uparrow$} 1.74\%)  & 0.763 (\textcolor{green}{$\uparrow$} 76.62\%)  & 0.756 (\textcolor{green}{$\uparrow$} 2.86\%)   & 0.764 (\textcolor{green}{$\uparrow$} 12.35\%)  & 0.763 (\textcolor{green}{$\uparrow$} 10.26\%)  & 0.752 (\textcolor{green}{$\uparrow$} 5.17\%)   \\
Cow           & \textbf{0.808 ± 0.130} & 0.764 (\textcolor{green}{$\uparrow$} 13.19\%) & 0.745 (\textcolor{green}{$\uparrow$} 132.09\%) & 0.768 (\textcolor{green}{$\uparrow$} 16.36\%)  & 0.762 (\textcolor{green}{$\uparrow$} 80.14\%)  & 0.752 (\textcolor{green}{$\uparrow$} 119.24\%) & 0.768 (\textcolor{green}{$\uparrow$} 18.89\%)  \\
Dog           & \textbf{0.848 ± 0.102} & 0.83 (\textcolor{green}{$\uparrow$} 11.86\%)  & 0.821 (\textcolor{green}{$\uparrow$} 88.30\%)  & 0.828 (\textcolor{green}{$\uparrow$} 6.15\%)   & 0.83 (\textcolor{green}{$\uparrow$} 52.57\%)   & 0.831 (\textcolor{green}{$\uparrow$} 58.89\%)  & 0.829 (\textcolor{green}{$\uparrow$} 11.28\%)  \\
Tie           & \textbf{0.700 ± 0.276} & 0.666 (\textcolor{green}{$\uparrow$} 6.22\%)  & 0.664 (\textcolor{green}{$\uparrow$} 80.43\%)  & 0.668 (\textcolor{green}{$\uparrow$} 2.93\%)   & 0.666 (\textcolor{green}{$\uparrow$} 17.05\%)  & 0.663 (\textcolor{green}{$\uparrow$} 22.32\%)  & 0.666 (\textcolor{green}{$\uparrow$} 3.10\%)   \\
Stop sign     & \textbf{0.886 ± 0.132} & 0.876 (\textcolor{green}{$\uparrow$} 4.41\%)  & 0.869 (\textcolor{green}{$\uparrow$} 58.00\%)  & 0.875 (\textcolor{green}{$\uparrow$} 3.18\%)   & 0.875 (\textcolor{green}{$\uparrow$} 13.20\%)  & 0.877 (\textcolor{green}{$\uparrow$} 20.80\%)  & 0.879 (\textcolor{green}{$\uparrow$} 37.34\%)  \\
Dolphin above & \textbf{0.732 ± 0.100} & 0.708 (\textcolor{green}{$\uparrow$} 10.28\%) & 0.7 (\textcolor{green}{$\uparrow$} 48.31\%)    & 0.706 (\textcolor{green}{$\uparrow$} 7.79\%)   & 0.708 (\textcolor{green}{$\uparrow$} 28.03\%)  & 0.707 (\textcolor{green}{$\uparrow$} 30.20\%)  & 0.693 (\textcolor{green}{$\uparrow$} 4.84\%)   \\
Dolphin below & \textbf{0.831 ± 0.075} & 0.775 (\textcolor{green}{$\uparrow$} 15.67\%) & 0.781 (\textcolor{green}{$\uparrow$} 99.74\%)  & 0.771 (\textcolor{green}{$\uparrow$} 7.98\%)   & 0.805 (\textcolor{green}{$\uparrow$} 70.19\%)  & 0.806 (\textcolor{green}{$\uparrow$} 87.44\%)  & 0.78 (\textcolor{green}{$\uparrow$} 14.54\%)   \\
Polyp         & \textbf{0.794 ± 0.145} & 0.771 (\textcolor{green}{$\uparrow$} 3.21\%)  & 0.763 (\textcolor{green}{$\uparrow$} 39.49\%)  & 0.77 (\textcolor{green}{$\uparrow$} 1.72\%)    & 0.771 (\textcolor{green}{$\uparrow$} 21.61\%)  & 0.772 (\textcolor{green}{$\uparrow$} 86.02\%)  & 0.769 (\textcolor{green}{$\uparrow$} 20.72\%)  \\
Skin          & \textbf{0.593 ± 0.195}          & 0.583 (\textcolor{red}{$\downarrow$} -1.69\%) & 0.568 (\textcolor{green}{$\uparrow$} 51.47\%)  & 0.565 (\textcolor{red}{$\downarrow$} -9.74\%)  & 0.578 (\textcolor{green}{$\uparrow$} 27.88\%)  & 0.586 (\textcolor{green}{$\uparrow$} 48.35\%)  & 0.592 (\textcolor{green}{$\uparrow$} 14.95\%)  \\
Breast        & \textbf{0.750 ± 0.126} & 0.723 (\textcolor{green}{$\uparrow$} 16.43\%) & 0.708 (\textcolor{green}{$\uparrow$} 61.64\%)  & 0.724 (\textcolor{green}{$\uparrow$} 7.42\%)   & 0.714 (\textcolor{green}{$\uparrow$} 26.15\%)  & 0.716 (\textcolor{green}{$\uparrow$} 34.59\%)  & 0.714 (\textcolor{green}{$\uparrow$} 20.61\%)  \\
Salt dome     & \textbf{0.844 ± 0.096} & 0.487 (\textcolor{red}{$\downarrow$} -5.07\%) & 0.48 (\textcolor{green}{$\uparrow$} 75.82\%)   & 0.488 (\textcolor{red}{$\downarrow$} -17.01\%) & 0.482 (\textcolor{green}{$\uparrow$} 38.90\%)  & 0.48 (\textcolor{green}{$\uparrow$} 56.86\%)   & 0.489 (\textcolor{red}{$\downarrow$} -13.30\%) \\
Chalk group   & \textbf{0.714 ± 0.101} & 0.418 (\textcolor{green}{$\uparrow$} 2.20\%)  & 0.428 (\textcolor{green}{$\uparrow$} 80.59\%)  & 0.426 (\textcolor{red}{$\downarrow$} -3.40\%)  & 0.423 (\textcolor{green}{$\uparrow$} 37.34\%)  & 0.42 (\textcolor{green}{$\uparrow$} 40.47\%)   & 0.418 (\textcolor{red}{$\downarrow$} -1.65\%) \\
\hline
\end{tabular}%
}
\end{table*}

\begin{table*}[ht]
\caption{\textbf{Prompt Encoder Finetuning Results:} Resulting mIoU (± standard deviation) across image datasets for automated and human strategies with Percent Change from Table I}
\label{tab:finetune-with-percentage-change}
\resizebox{\textwidth}{!}{%
\begin{tabular}{cccccccc}
\hline
\textbf{Category} & \textbf{SAM finetuned on} & \textbf{SAM finetuned on} & \textbf{SAM finetuned on} & \textbf{SAM finetuned on}  & \textbf{SAM finetuned on} & \textbf{SAM finetuned on} & \textbf{SAM finetuned on} \\
 & \textbf{Human prompts}    & \textbf{Random prompts} & \textbf{Saliency prompts} & \textbf{K-Medoids prompts}  & \textbf{Entropy prompts} & \textbf{Maximum Dist. prompts} & \textbf{Shi-Tomasi prompts} \\
\hline
\hline
Baseball bat  & 0.729 (\textcolor{red}{$\downarrow$} -2.41\%)   & 0.684 ( 0.0\%)   & 0.509 (\textcolor{green}{$\uparrow$} 20.62\%)  & \textbf{0.733} (\textcolor{green}{$\uparrow$} 1.24\%)   & 0.690 (\textcolor{green}{$\uparrow$} 5.66\%)   & 0.644 (\textcolor{green}{$\uparrow$} 1.90\%)   & 0.690 \textcolor{red}{$\downarrow$} -1.57\%)  \\
Bird          & 0.704 (\textcolor{green}{$\uparrow$} 4.0\%)   & 0.723 (\textcolor{green}{$\uparrow$} 17.56\%)  & 0.477 (\textcolor{green}{$\uparrow$} 54.87\%)  & \textbf{0.725} (\textcolor{green}{$\uparrow$} 12.4\%)    & 0.662 (\textcolor{green}{$\uparrow$} 37.02\%)  & 0.590 (\textcolor{green}{$\uparrow$} 29.39\%)  & 0.709 (\textcolor{green}{$\uparrow$} 14.35\%)  \\
Bus           & \textbf{0.900} (\textcolor{green}{$\uparrow$} 12.08\%)   & 0.873 (\textcolor{green}{$\uparrow$} 47.06\%)  & 0.672 (\textcolor{green}{$\uparrow$} 325.32\%)  & 0.889 (\textcolor{green}{$\uparrow$} 39.81\%)  & 0.740 (\textcolor{green}{$\uparrow$} 106.14\%)  & 0.692 (\textcolor{green}{$\uparrow$} 139.79\%)  & 0.868 (\textcolor{green}{$\uparrow$} 58.39\%)  \\
Cat           & \textbf{0.904} (\textcolor{green}{$\uparrow$} 1.91\%)   & 0.876 (\textcolor{green}{$\uparrow$} 13.61\%)  & 0.775 (\textcolor{green}{$\uparrow$} 59.14\%)  & \textbf{0.904} (\textcolor{green}{$\uparrow$} 9.58\%)   & 0.883 (\textcolor{green}{$\uparrow$} 51.29\%)  & 0.844 (\textcolor{green}{$\uparrow$} 66.14\%)  & 0.891 (\textcolor{green}{$\uparrow$} 12.07\%)  \\
Clock         & 0.797 (\textcolor{red}{$\downarrow$} -2.09\%)  & \textbf{0.833} (\textcolor{green}{$\uparrow$} 11.79\%)  & 0.578 (\textcolor{green}{$\uparrow$} 33.79\%)  & 0.832 (\textcolor{green}{$\uparrow$} 13.15\%)  & 0.759 (\textcolor{green}{$\uparrow$} 11.61\%)  & 0.759 (\textcolor{green}{$\uparrow$} 9.68\%)   & 0.825 (\textcolor{green}{$\uparrow$} 15.35\%)  \\
Cow           & 0.832 (\textcolor{green}{$\uparrow$} 2.97\%)  & \textbf{0.840} (\textcolor{green}{$\uparrow$} 24.44\%)  & 0.618 (\textcolor{green}{$\uparrow$} 92.53\%)  & 0.836 (\textcolor{green}{$\uparrow$} 26.67\%)  & 0.779 (\textcolor{green}{$\uparrow$} 84.14\%)  & 0.765 (\textcolor{green}{$\uparrow$} 123.30\%)  & 0.823 (\textcolor{green}{$\uparrow$} 27.42\%)  \\
Dog           & 0.837 (\textcolor{red}{$\downarrow$} -1.29\%)  & 0.830 (\textcolor{green}{$\uparrow$} 11.83\%)  & 0.612 (\textcolor{green}{$\uparrow$} 40.32\%)  & \textbf{0.868} (\textcolor{green}{$\uparrow$} 11.28\%)   & 0.789 (\textcolor{green}{$\uparrow$} 45.07\%)  & 0.745 (\textcolor{green}{$\uparrow$} 42.41\%)  & 0.843 (\textcolor{green}{$\uparrow$} 13.15\%)  \\
Tie           & 0.724 (\textcolor{green}{$\uparrow$} 3.43\%)  & 0.691 (\textcolor{green}{$\uparrow$} 10.21\%)  & 0.479 (\textcolor{green}{$\uparrow$} 30.16\%)  & \textbf{0.748} (\textcolor{green}{$\uparrow$} 15.25\%)  & 0.648 (\textcolor{green}{$\uparrow$} 13.88\%)  & 0.599 (\textcolor{green}{$\uparrow$} 10.52\%)  & 0.699 (\textcolor{green}{$\uparrow$} 8.20\%)   \\
Stop sign     & 0.888 (\textcolor{green}{$\uparrow$} 0.22\%)  & 0.886 (\textcolor{green}{$\uparrow$} 5.60\%)   & 0.674 (\textcolor{green}{$\uparrow$} 22.55\%)  & \textbf{0.894} (\textcolor{green}{$\uparrow$} 5.42\%)   & 0.854 (\textcolor{green}{$\uparrow$} 10.46\%)  & 0.837 (\textcolor{green}{$\uparrow$} 15.28\%)  & 0.890 (\textcolor{green}{$\uparrow$} 39.06\%)  \\
Dolphin above & 0.754 (\textcolor{green}{$\uparrow$} 3.01\%)  & 0.751 (\textcolor{green}{$\uparrow$} 16.96\%)  & 0.722 (\textcolor{green}{$\uparrow$} 53.05\%)  & 0.743 (\textcolor{green}{$\uparrow$} 13.41\%)  & 0.724 (\textcolor{green}{$\uparrow$} 30.56\%)  & 0.708 (\textcolor{green}{$\uparrow$} 30.20\%)  & \textbf{0.761} (\textcolor{green}{$\uparrow$} 15.11\%)  \\
Dolphin below & \textbf{0.878} (\textcolor{green}{$\uparrow$} 5.65\%)  & 0.825 (\textcolor{green}{$\uparrow$} 23.13\%)  & 0.687 (\textcolor{green}{$\uparrow$} 75.70\%)  & 0.847 (\textcolor{green}{$\uparrow$} 18.62\%)  & 0.808 (\textcolor{green}{$\uparrow$} 70.83\%)  & 0.813 (\textcolor{green}{$\uparrow$} 89.07\%)  & 0.849 (\textcolor{green}{$\uparrow$} 24.65\%)  \\
Polyp         & 0.843 (\textcolor{green}{$\uparrow$} 6.16\%)  & 0.856 (\textcolor{green}{$\uparrow$} 14.59\%)  & 0.712 (\textcolor{green}{$\uparrow$} 30.15\%)  & \textbf{0.871} (\textcolor{green}{$\uparrow$} 15.06\%)  & 0.781 (\textcolor{green}{$\uparrow$} 23.21\%)  & 0.719 (\textcolor{green}{$\uparrow$} 73.49\%)  & 0.848 (\textcolor{green}{$\uparrow$} 33.16\%)  \\
Skin          & 0.784 (\textcolor{green}{$\uparrow$} 32.20\%)  & 0.821 (\textcolor{green}{$\uparrow$} 38.41\%)  & 0.683 (\textcolor{green}{$\uparrow$} 82.13\%)  & \textbf{0.859} (\textcolor{green}{$\uparrow$} 37.13\%)  & 0.792 (\textcolor{green}{$\uparrow$} 28.13\%)  & 0.799 (\textcolor{green}{$\uparrow$} 42.53\%)  & 0.785 (\textcolor{green}{$\uparrow$} 52.43\%)  \\
Breast        & 0.804 (\textcolor{green}{$\uparrow$} 7.20\%)  & 0.783 (\textcolor{green}{$\uparrow$} 26.04\%)  & 0.649 (\textcolor{green}{$\uparrow$} 48.18\%)  & \textbf{0.831} (\textcolor{green}{$\uparrow$} 23.28\%)  & 0.762 (\textcolor{green}{$\uparrow$} 34.61\%)  & 0.733 (\textcolor{green}{$\uparrow$} 37.84\%)  & 0.807 (\textcolor{green}{$\uparrow$} 36.15\%)  \\
Salt dome     & 0.831 (\textcolor{red}{$\downarrow$} -1.54\%)  & 0.809 (\textcolor{green}{$\uparrow$} 57.7\%)  & 0.499 (\textcolor{green}{$\uparrow$} 82.78\%)  & \textbf{0.881} (\textcolor{green}{$\uparrow$} 49.91\%)  & 0.638 (\textcolor{green}{$\uparrow$} 83.84\%)  & 0.711 (\textcolor{green}{$\uparrow$} 132.32\%)  & 0.820 (\textcolor{green}{$\uparrow$} 45.39\%)  \\
Chalk group   & 0.695 (\textcolor{red}{$\downarrow$} -2.66\%)  & 0.617 (\textcolor{green}{$\uparrow$} 50.86\%)  & 0.332 (\textcolor{green}{$\uparrow$} 40.09\%)  & \textbf{0.735} (\textcolor{green}{$\uparrow$} 66.67\%)  & 0.523 (\textcolor{green}{$\uparrow$} 69.81\%)  & 0.612 (\textcolor{green}{$\uparrow$} 104.67\%)  & 0.671 (\textcolor{green}{$\uparrow$} 57.88\%)  \\
\hline       
\end{tabular}%
}
\end{table*}

Results for this experiment are detailed in Table \ref{tab:human-inclusion-arrows}, where the mIoUs for each image dataset across the distinct exclusion sampling strategies is shown. The percent change from Table \ref{tab:strategies-best} is also indicated. Unlike the results from using human exclusion points in Table \ref{tab:human-exclusion2}, Table \ref{tab:human-inclusion-arrows} instead shows widespread mIoU score improvement across all image datasets and sampling strategies when utilizing human inclusion points. Specifically, utilizing human inclusion points results in automated strategies experiencing a $36.3\%$ increase in mIoU scores when compared to results from Table \ref{tab:strategies-best} on average.
Hence, Table \ref{tab:human-inclusion-arrows} demonstrates that performance across all strategies for most image datasets becomes closer to human mIoU performance, despite different exclusion points being used. Overall, these results indicate that inclusion points tend to impact segmentation performance more than exclusion points. However, the datasets from the seismic domain still display a large difference between human mIoU scores and scores attained from using a combination of human inclusion points and exclusion points generated from automated strategies. This suggests that certain domains seem to be more sensitive to the placement of both inclusion and exclusion points, which only human annotators were able to reconcile.

\subsection{Finetuning}

We then evaluate SAM's adaptation capabilities to handle a variety of prompting strategies, both human and automated, through finetuning experiments. Finetuning results are evaluated on test sets for each of the 16 image datasets and reported in Table \ref{tab:finetune-with-percentage-change}. We observe here that the gap is largely closed, with automated methods achieving performance improvements anywhere from $22\%$ to $68\%$ on average when compared to Table \ref{tab:strategies-best}. These improvements are particularly prevalent with the K-Medoids automated approach. When finetuning with prompts generated via K-Medoids, we observe that this approach surpasses all other strategies (including human performance) on 11 out of the total 16 datasets. Similarly, finetuning with prompts generated by the Random strategy produce the highest segmentation scores in 2 out of the 16 datasets. Thus, we see significant performance benefits when finetuning with automated strategies.

Additionally, finetuning on the Saliency and Maximum Distance prompts generates the highest percent change with respect to Table \ref{tab:strategies-best}. For instance, finetuning on the Saliency strategy produces an average $68.21\%$ percent increase from the results for this strategy shown in Table \ref{tab:strategies-best}. We see that the finetuning results for this strategy in Table \ref{tab:finetune-with-percentage-change} can sometimes produce comparable results to humans and other automated strategies, shown in datasets like Skin and Dolphin below. These results indicate that even poor-performing strategies (as indicated by Table \ref{tab:strategies-best}) can be utilized in a finetuning manner to provide increased segmentation scores.

Overall, we observe that \emph{finetuning the prompt encoder is an effective way to adapt SAM to individual prompting strategies}, both for in-distribution and OOD domains. It is worth noting, however, that the main drawback in this approach lies in its computational cost, since even finetuning the prompt encoder component of a foundation model like SAM is resource-intensive due to the cost of running inference on the entire model iteratively.


\subsection{Performance decoding}

\begin{figure*}
    \centering
    \includegraphics[width=.95\linewidth]{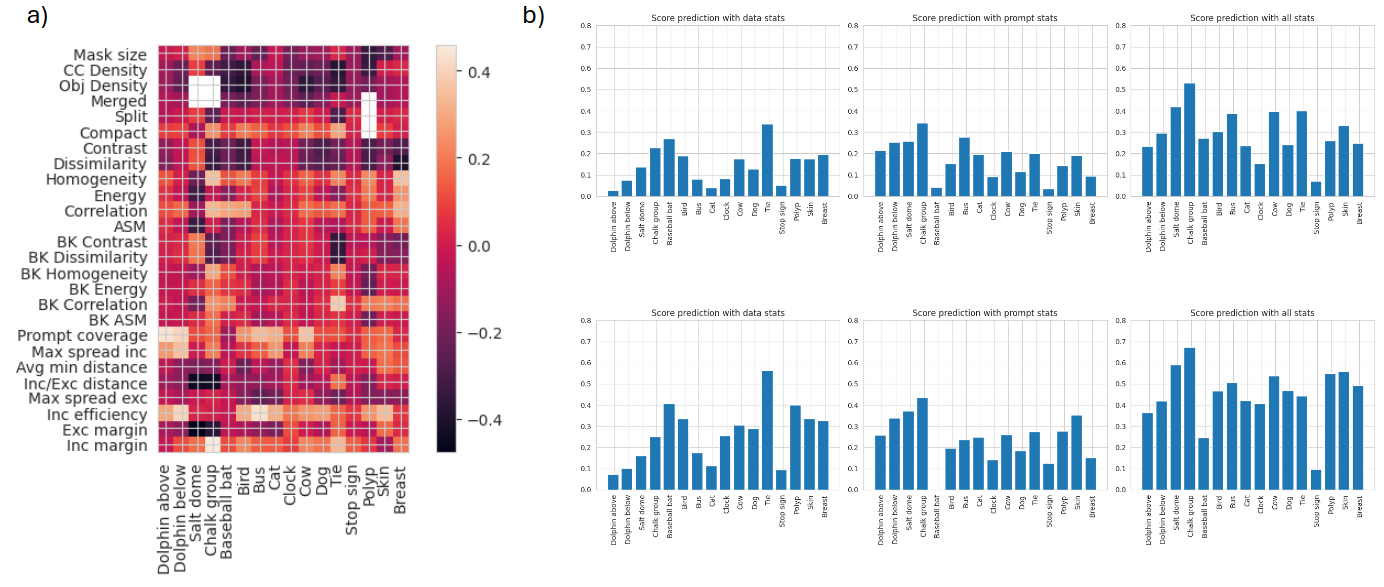}
    \caption{(b): $R^2$ prediction scores across individual datasets. Top: prediction score using linear regression. Bottom: prediction score using a 2nd-degree polynomial regression. From left to right: prediction using only data features, only prompt features, or both sets of features}
    \label{fig:decoding}
\end{figure*}

\begin{figure}
    \centering
    \includegraphics[width=0.95\linewidth]{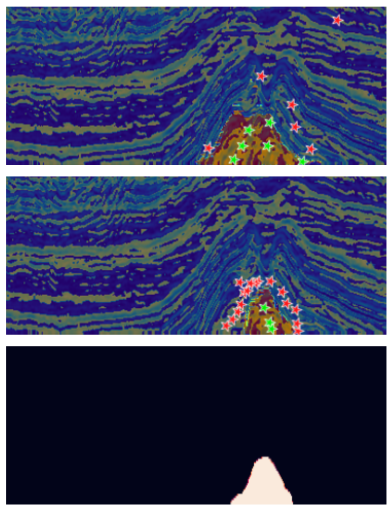}
    \caption{Visual example of two different strategies in seismic (salt dome) prompting. Top: example strategy with high exclusion margin and high inclusion max spread. Middle: example strategy with low exclusion margin and low inclusion max spread. Bottom: Segmentation ground truth mask. }
    \label{fig:seismic}
\end{figure}

\subsubsection{Correlation analysis}
To measure the role our curated data and prompt-level features play into segmentation performance, we evaluate the correlation between each of them and the IoU score for all the gathered data using the Pearson correlation coefficient. We show these results in Figure \ref{fig:decoding}.a, where the vertical axis corresponds to the features and the horizontal axis corresponds to a dataset, and an entry in position $(i,j)$ in the matrix represents the correlation coefficient between the $i$-th feature and the IoU score for the $j$-th dataset. The maximum correlation value attained is 0.45, and the minimum one is -0.47, which tells us for instance that higher inclusion margin and lower exclusion margin are beneficial to performance in the chalk group and salt dome seismic datasets, respectively. While in some cases we can see reasonable correlations in individual feature-dataset pairs, it is generally apparent that these features do not account for segmentation performance on their own. Thus at this level, we can see that no single data or prompt-level feature is directly related to the score across all datasets, which suggests that \textit{there might not be a one-size-fits-all prompting strategy that does well across every type of dataset}. 

\subsubsection{Decoding}

While the previous analysis illustrates the existence of some degree of relationship between features and performance, it is constrained to \emph{pairwise linear} relationships, and it is likely that these features influence segmentation performance in a more complex manner. Figure \ref{fig:decoding}.b explores this possibility: we attempt to predict the performance by performing linear (top) and quadratic (bottom) regression using either the data (left), prompt (middle) or both (right) sets of features. As we can see in the quadratic decoding experiments, an $R^2$ score greater than 0.5 is achieved for many of the image categories, suggesting that the interplay between these features is indicative of segmentation performance in many cases. In particular, we can see that for domains that are OOD for SAM (seismic and medical data), the features are highly predictive of performance. This allows us to then analyze how each feature contributes to performance and to what extent, by looking at the regression coefficients. For instance, the most determinant features for salt dome segmentation quality appear to be inclusion max spread with a coefficient of -0.22 and exclusion margin with a coefficient of -0.14, suggesting that an effective strategy for this type of data might be closely prompting on the boundary with exclusion points, while avoiding covering a large part of the salt dome with inclusion prompts. This aligns with our empirical observations, of which we show an example in Figure \ref{fig:seismic}: the set of prompts that most closely aligns with the above described strategy (middle figure), achieves the best segmentation result. This type of analysis provides a \emph{simple yet interpretable way of designing effective prompting strategies for out-of-distribution domains}.

\section{Discussion and Conclusion}

We have leveraged the PointPrompt dataset to address the current limitations in understanding the difference between human and automated prompting, identifying and measuring the gap between them. We have also studied the role inclusion and exclusion points play in this difference, and identified that inclusion points contribute to a much larger degree. Furthermore, we find that the performance of automated prompting methods is significantly improved by finetuning SAM's prompt encoder to a given prompting approach. It is worth noting, however, that this improvement is both data and computationally costly, since it involves generating the image/prompt pairs and training the prompt encoder (while the rest of SAM's components are kept frozen, even running inference on them to calculate the loss is resource-intensive).


We have also identified a series of intuitive features to facilitate the study of both human and automated prompting strategies. Our performance decoding analysis not only highlights the value of human prompting strategies, but also sheds light on key factors that determine segmentation performance in the OOD domains we have analyzed, and allows for domain specific prompt strategy design. Overall, these benchmark tasks along with the PointPrompt dataset constitute a step forward in defining what the computer vision community can do in the design of effective prompts, whether that be from a finetuning or feature-based perspective. As datasets like this expand and become more commonplace, human-driven insights into effective prompts for various domains will further accelerate research in prompt design and transferability. 

\section*{Acknowledgments}
This work is partially supported by the ML4Seismic Industry Partners
at Georgia Tech. This material is also based upon work supported by the National Science Foundation Graduate Research Fellowship under Grant No. DGE-2039655. Any opinion, findings, and conclusions or recommendations expressed in this material are those
of the authors(s) and do not necessarily reflect the views of the National Science Foundation.

\bibliography{ref}

\end{document}